\pdfoutput=1
\documentclass[11pt]{article}

\usepackage{coling}
\usepackage{graphicx}

\usepackage[utf8]{inputenc}
\usepackage[T1]{fontenc}
\usepackage{multirow}
\usepackage{caption}
\usepackage{subcaption}
\usepackage{mdframed}
\usepackage[most]{tcolorbox}

\newcommand{\codeRepo}{\url{https://github.com/wendli01/legal_analysis_llm}}

\newcommand{\prompt}[3]{
\begin{figure}
    \centering
        \begin{minipage}{0.9\linewidth}
    \small
    \begin{tcolorbox}{}
    #1
    \end{tcolorbox}
  \end{minipage}
    \caption{#2}
    \label{#3}
\end{figure}
}

    \title{On the Suitability of pre-trained foundational LLMs for Analysis in German Legal Education }

    \author{Lorenz Wendlinger\textsuperscript{1},
     \textbf{Christian Braun\textsuperscript{2}},
    \textbf{Abdullah Al Zubaer\textsuperscript{1}},
    \\
    \textbf{Simon Alexander Nonn\textsuperscript{2}},
    \textbf{Sarah Großkopf\textsuperscript{2}},
    \textbf{Christofer Fellicious\textsuperscript{1}},
    \\
    \textbf{Michael Granitzer\textsuperscript{1}}
    \\
    \\
    \textsuperscript{1}Chair of Data Science,
    Universität Passau, Germany\\
    \textsuperscript{2}Institut für Rechtsdidaktik,
    Universität Passau, Germany
    \\
\small{
    \textbf{Correspondence:} \href{mailto:lorenz.wendlinger@uni-passau.de}{lorenz.wendlinger@uni-passau.de}
  }
  \\
}

\begin{document}

\maketitle


    \begin{abstract}
        We show that current open-source foundational LLMs possess instruction capability and German legal background knowledge that is sufficient for some legal analysis in an educational context.
        However, model capability breaks down in very specific tasks, such as the classification of ``Gutachtenstil'' appraisal style components, or with complex contexts, such as complete legal opinions.
        Even with extended context and effective prompting strategies, they cannot match the Bag-of-Words baseline.
        To combat this, we introduce a Retrieval Augmented Generation based prompt example selection method that substantially improves predictions in high data availability scenarios.
        We further evaluate the performance of pre-trained LLMs on two standard tasks for argument mining and automated essay scoring and find it to be more adequate.
        Throughout, pre-trained LLMs improve upon the baseline in scenarios with little or no labeled data with Chain-of-Thought prompting further helping in the zero-shot case. 
    \end{abstract}


    \section{Introduction}\label{sec:introduction}

    Large language models (LLMs) are trained on web-scale corpora that encompass a significant cross-section of human knowledge and communication.
    They offer generalization abilities that allow natural language instruction without any or with little need for fine-tuning.
    With \citeposs{ainslie2023gqa} Grouped-Query Attention, context windows are large enough to provide exhaustive and relevant background information.
    For this reason, they are poised to be a replacement for task-specific models in niche and evolving domains.

    Legal analysis is such a domain, with a large quantity of new laws and court decisions that make human analysis time consuming, while the structure and style of legal appraisal remain largely static.
    We want to empirically verify the assumption that the current crop of open-source local LLMs can understand this appraisal style by checking if they can correctly recognize it.

    The German \textit{Gutachtenstil} (appraisal style) is similar in structure and purpose to the IRAC (\textit{Issue, Rule, Application, Conclusion}) method of legal analysis. It starts with a \textit{Major claim} that names the legal question that is analyzed, the \textit{Definition} that defines the requirements that need to be fulfilled to affirm the major claim. After that, in the \textit{Subsumtion}, the facts of the case are applied to the different elements of the definition. Lastly, the \textit{Conclusion} is reached by answering the question raised in the major claim.
    It is intended as a guide to help structure the thinking of law students.
    In German law, this schema is complex enough to be a focus of entry and intermediate level college lectures, with some universities even offering dedicated courses to teach the appraisal style.
    As it is seldom found in legal practice outside of universities, but rather substituted with the modified \textit{Urteilsstil}, that starts with the conclusion and then uses the steps of definition and subsumption to justify the result, legal corpora may not contain enough examples to implicitly encode this information.

    We consider this to be the minimum level of evidence required as proof for legal understanding.
    If a legal practitioner cannot appropriately structure the constituents of arguments into this appraisal style, they show a lack of understanding which may also result in content-specific mistakes.
    Similarly, if any student cannot recognize the elements of the appraisal style and fails to apply it in their own reasoning, they run a risk of exhibiting lapses in logic, of overlooking facts or legal requirements.
    Therefore, we start our analysis by testing whether LLMs can reliably identify the elements of this argumentation style.

    To test the upper bounds of LLM capability, we perform a specific test to check the assessment capabilities of the models in the complex but controlled context of a challenging new dataset.
    The task of grading full legal opinions requires close to complete legal understanding.
    It can only be facilitated with a large catalog of background knowledge, comprised of the legal context, concurring and dissenting legal opinions and secondary literature.
    The examination of multiple statutes and constituent facts also necessitates the ability to appropriately contextualize and structure the components of the legal network that is established around an appraisal.
    For this purpose, we introduce a dataset of full legal papers written and graded in the context of legal education.

    We further investigate the performance of pre-trained LLMs in two simpler related tasks in the realm of argument mining and automated essay scoring.
    This provides some context on the influence of task difficulty, language barriers, and domain-specific adaptations.

    The used codebase as well as code for all experimental setups with results are available at \codeRepo.

    \section{Related Work}\label{sec:related-work}

    Recent work explores using generative LLMs for argument mining and essay scoring tasks, though most methods focus on extractive models instead.
    We summarize relevant results from automated essay scoring and argument mining below.

    \citet{rodriguez2019language} apply two early transformer models to the Automatic Essay Scoring (AES) dataset from \cite{taghipour2016neural}.
    They find that these models are competitive with Long Short-Term Memory Networks (LSTMs) and can outperform them as well as human scoring in this task if combined into a heterogeneous ensemble. \citet{yang2020enhancing} use multiple losses to fine-tune language models on the Automated Student Assessment Prize (ASAP) dataset. \citet{ormerod2021automated} use efficient Transformer variants to score essays. They find that they can achieve equivalent performance at a significant efficiency increase.
    
    \citet{lai2023large} provide a comprehensive overview of LLMs in Chinese legal applications.
    They find that there are multiple models that are fine-tuned on Chinese legal QA corpora as well as more general models for legal consultation and reasoning.
    They point out the major problems with legal LLMs, mainly the difficulty of data quality control as well as bias and interpretability issues.
    They further allude to the inherent uncertainty of legal concepts, that makes the legal domain challenging to adapt LLMs to.

    \citet{mizumoto2023exploring} explore the applicability of \texttt{GPT-3.5} to AES on the TOEFL11 corpus.
    They find that it generally lags behind the baseline Bayesian regression model operating on linguistic features only, but that combining the two leads to a slight improvement.

    \citet{colombo2024saullm} fine-tune Mistral 7B on a large English legal corpus and publish the resulting \texttt{SaulLM-7B}.
    They perform general instruction fine-tuning followed by specific legal instruction fine-tuning with synthetic data generated by the base model.
    They find that the resulting model outperforms the base in the newly introduced LegalBench-Instruct, Legal-MMLU and perplexity, i.e. quality-of-fit, on legal documents.

    \citet{xiao2024automation} experiment with using \texttt{GPT} models for automated essay scoring and semi-supervised grading systems.
    They find that a fine-tuned variant of \texttt{GPT-3.5} outperforms other models and the traditional baseline in most ASAP sets as well as on a private Chinese student English essay dataset.

    \citet{stahl2024exploring} explore the impact of different prompts and prompting strategies in automated essay scoring.
    They find that the specific prompt has a measurable effect, while more complex strategies can be detrimental.

    The use of LLMs for argument mining has seen growing interest in recent times. For instance, \citet{10.3389/frai.2023.1278796} analyze the performance of \texttt{GPT-3.5} and \texttt{GPT-4} for argument component classification via prompting in the legal domain.
    They find that open-source local models perform better than propriety models and demonstrate the significance of including similar examples (in contrast to random and dissimilar) examples in prompts to boost the model's performance. 
    
    \citet{otiefy-alhamzeh-2024-exploring-large} explore another sub-component of argument mining in relation detection between argument components using fine-tuned transformer-based models and \texttt{GPT-4} for the financial domain. 
    The find \texttt{GPT-4}'s performance to be better than the fine-tuned transformer-based models.

    Similarly, \citet{gorur2024largelanguagemodelsperform} demonstrate the effectiveness of \texttt{Llama 2} and \texttt{Mixtral} for argument relation identification where given a pair of argument components, the model classifies the relation between them (support or attack). 
    They find that \texttt{Llama 2} and \texttt{Mixtral} outperformed traditional RoBERTa-based baseline in different domains ranging from essays to online debate forums.

    \section{Methods}\label{sec:methods}

    We use multiple prompt engineering techniques for both the design of our analysis systems and post-mortem analysis.
    These include forays into few-shot prompting, chain-of-thought prompting, retrieval augmented generation, instruction following and pseudonymization.


    We consider an excerpt (\autoref{tab:livebench}) of relevant categories from \citeposs{livebench} \textit{LiveBench}\footnote{\url{https://livebench.ai/}}, a contamination-free LLM leaderboard.    
    While \texttt{Llama 3} and \textit{GPT-3.5} are close in the overall score, they exhibit opposite and complementary strengths in the two most relevant tasks, reasoning and language comprehension.
    This allows for speculation as to which of the two skills is more relevant in legal analysis.
    \texttt{Mixtral 8x7B} is competitive in reasoning but falls short in the language comprehension task.

    \subsection{Generative Pre-trained Transformer}

    The \texttt{GPT} family of LLMs is based on a decoder-only architecture that is trained on a web-scale corpus of documents and made available for commercial use by OpenAI.
    \texttt{GPT-3.5} is a model that is optimized for assistant tasks.

    \subsection{Llama 3}

    \citeposs{llama3modelcard} Llama 3 is an improved decoder-only transformer with a more advanced tokenizer using \citeposs{ainslie2023gqa} grouped query attention. It is trained using a context window of size 8,192 tokens and boasts an effective inference maximum of approx. 128k.
    We use the \textit{Meta-Llama-3-70B-Instruct} checkpoint \footnote{\url{https://huggingface.co/meta-llama/Meta-Llama-3-70B-Instruct}}.
    
    \subsection{Mixtral}

    \texttt{Mixtral} \cite{mixtralmodelcard} is a \textbf{S}parse \textbf{M}ixture-\textbf{o}f-\textbf{E}xperts (SMoE) model built from 8 Mistral-7B experts and purported to be on par with or outscore \texttt{GPT-3.5} as well as \texttt{Llama2 70B} in several language understanding and LLM assistant tasks.
    It boasts a context length of 32k tokens through local attention and efficient sparse inference via a router network that selects 2 out of 8 experts for each layer and token.
    It exhibits low bias and hallucination tendencies while showing fluency in, among other languages, German.
    We make use of the \texttt{Mixtral-8x7B-Instruct-v0.1} checkpoint \footnote{\url{https://huggingface.co/mistralai/Mixtral-8x7B-Instruct-v0.1}}.

    \subsection{Jina Embeddings}
     \citet{mohr2024multi} propose the more focused and lightweight BERT-based Jina Embedding model.
    We use the German-English bilingual 161 million \texttt{base-de} version\footnote{\url{https://huggingface.co/jinaai/jina-embeddings-v2-base-de}} to generate embeddings for Retrieval augmented Generation.

    \subsection{Prompting Strategies}

    We use and describe here several prompting strategies from the literature, of which \citet{sahoo2024systematic} provide a systematic survey, with slight adaptations to our domain.

    In \citeposs{radford2019language} zero-shot prompting, a model is only instructed with the task plus context and background knowledge, if available.
    This requires no labeled training data and is therefore compelling for novel and niche applications.

    In contrast, \citet{brown2020language} show that integrating even a few examples can enhance model responses.
    This adaptation is quantitatively cheap but requires carefully chosen high quality data.
    The added complexity also results in longer prompts and runtimes as well as a higher likelihood of unintended bias.

    \citet{lewis2020retrieval} use Retrieval Augmented Generation (RAG) to provide additional context based on a knowledge base, e.g. the Wikipedia.
    They use a pre-trained encoder to obtain dense representations for all snippets in the knowledge base and integrate those that have maximum inner-product similarity with the query embedding into the prompt.
    
    \citet{wei2022chain} show that allowing LLMs to explain their decision step-by-step is beneficial in many complex tasks.
    We find that this Chain-of-Thought prompting impacts instruction following, as it reduces the importance of the initial instruction.
    For this reason, we employ pattern matching to find the most mentioned category in the model response and report this as the result.

    \citet{zhang2022automatic} propose \textit{Auto-CoT}  to bridge the gap between Zero-Shot-CoT and Few-Shot-CoT with manual examples.
    They form clusters of similar prompts and for those generate rationales that are used as prompt examples.
    These are pre-selected using a criterion geared towards simplicity.

    Similar to this work, we propose Generated Artificial Rationales (GAR) that are applicable to classification and regression scenarios with moderate to large amounts of labeled training data without rationales.    
    We slightly adapt \citeposs{zhang2022automatic} selection criterion to ensure the correctness of the rationale that was generated under the same conditions that the model operates under in testing.
    As a rejection method this is inefficient, but we can integrate a similar sampling technique to Auto-CoT and thereby include a relevancy bias.

We choose to develop simple prompts in lieu of complex roles or instruction as \cite{stahl2024exploring} shows that their effectiveness can be very specific and task dependent.
However we make use of the instruction fine-tuning of the investigated models by providing a separate system prompt (c.f. prompt \ref{prompt:spwsle_sys}) and result prompt in the user role (c.f. prompt \ref{prompt:spwsle}).

    \section{Results}\label{sec:results}

We evaluate the suitability of pre-trained generative LLMs on 4 datasets across the two tasks of argument mining and essay scoring.
We first describe the datasets and accompanying evaluation methodology before reporting the main results as well as ablation studies.

    \subsection{Datasets}\label{subsec:datasets}

We evaluate the effectiveness of pre-trained generative LLMs on 4 datasets, with 2 each for argumentation mining and essay scoring, c.f. table \ref{tab:datasets} for their properties.
For both tasks, we included a simpler and standard dataset to investigate the impact of outright task difficulty.
    
    \begin{table}
        \footnotesize
        \begin{subtable}[h]{\linewidth}
        \centering
        \begin{tabular}{l||r|c|r|r}
             Dataset&  Class. & Docs & Items & Length  \\\hline\hline
             SPWSLE$^1$                    & 6    & 382   & 14734 & 627   \\\hline
             CIMT$^2$    CD\_C             & 4    & 366   & 1704  & 62   \\\hline
        \end{tabular}
        \caption{German Argument Mining}
        \end{subtable}

        \begin{subtable}[h]{\linewidth}
        \centering
        \begin{tabular}{l||l|r|c|r}
             Dataset&  Lang  &Points & Docs  & Length  \\\hline\hline
             GSHA                                           &de     & 0-18 & 71         & 5877 \\\hline
                         \hspace{8.5ex} 1                    &          &       & 1783    & 350    \\
            ASAP$^3$ 2     &en    &10-60 & 1800     & 350    \\
                         \hspace{8.5ex} 8                    &          &       & 723      & 650    \\
        \end{tabular}
        \caption{Automated Essay Scoring}
        \end{subtable}
        
        \caption{Dataset statistics with annotation schema, number of documents, number of annotated items and mean document length in tokens. Datasets from $^1$: \cite{weber2023structured}, $^2$: \cite{romberg2021citizen}, $^3$: \cite{hamner2012hewlett}.}
        \label{tab:datasets}
    \end{table}

    \subsubsection{SPWSLE}
    The Dataset introduced in \textit{\textbf{S}tructured \textbf{P}ersuasive \textbf{W}riting \textbf{S}upport in \textbf{L}egal \textbf{E}ducation: A Model and Tool for German Legal Case Solutions} \cite{weber2023structured} comprises 413 student submitted solutions to 4 different legal cases with a mean length of 60.8 sentences.
    We abbreviate the classes as follows: \textbf{MC} = Major Claim, \textbf{C} = Conclusion, \textbf{D} = Definition, \textbf{S} = Subsumption, \textbf{LC} = Legal Claim, \textbf{P} = Premise, \textbf{N} = None.

    This dataset splits the class Subsumption into two different subclasses, Premise and Legal Claim. The premise serves as a statement of the facts as they relate to the Definition, and the Legal Claim then matches the facts of the Premise to the Definition. 

    \subsubsection{Graded Strafrecht Hausarbeiten}

    We present here the \textbf{G}raded \textbf{S}trafrecht \textbf{H}aus\textbf{A}rbeiten dataset which is distinguished by its small size as well as the complexity of its samples and the task.
    It is intended as a transfer learning challenge for models with high instruction and language capability as it does not provided enough data for full fine-tuning on the main task.
    As such, it encompasses 76 distinct student solutions to a legal problem complex with a final grade on an 18 point scale attached to each.
    
The student solutions were created by intermediate and advanced German law students in the course of their regular studies. 
They were provided with the facts of a criminal law case and tasked with solving it over several weeks. Use of legal commentaries, textbooks and databases was allowed. 
Out of a larger cohort, 76 students then provided their consent for the use of their work and its grade for our research.
Afterwards, the works were corrected by research assistants on a scale from 0-18, with 18 being the best, and 4 being the minimum score required to pass.   

    \subsubsection{ASAP AES}

    The \textbf{A}utomated \textbf{S}tudent \textbf{A}ssessment \textbf{P}rize \cite{hamner2012hewlett} contains English student essays for 8 different tasks.
    They are assigned combined scores between 10 and 60 points each.
    Here we select sets 1,2 and 8 as they are all from the \textit{Persuasive/Narrative/Expository} category as per \cite{xiao2024automation}. 
    We focus on set 8 as it contains the longest and most complex documents, c.f. table \ref{tab:datasets}.

    \subsubsection{CIMT}
    \citeposs{romberg2021citizen} CIMT argument mining dataset consists of German language public feedback statements to 5 different infrastructure projects or concepts.
    These contributions are annotated on a sentence level with the categories \textit{Major Position}, \textit{Premise}, \textit{Major Position + Premise} or \textit{None}.
    It is therefore similar to SPWSLE, but with a simplified annotation schema and much shorter texts.

    \subsection{Evaluation Methodology}\label{subsec:evaluation-methodology}

    
    For all experiments, we report the performance of a linear SVM with Bag-of-Words features as an established baseline.

    We re-phrase the classification of appraisal style components as a joint task, that involves the recognition of both the subsumption component as well as its sub-components in one pass.
    This is in contrast to \cite{weber2023structured}, where the authors propose a two-tiered task that first extracts the main components and further sub-divides the subsumption.
    It is not exactly clear how they do this, however the presence of a \textit{None} category in both tiers in their evaluation coupled with the absence of a respective category in the annotations hints at a usage as a masking placeholder.
    Furthermore there is no mapping of annotation categories to classes included.
    Our estimate is the following: e$_1$=\textbf{M}ajor \textbf{C}laim, e$_2$=\textbf{C}onclusion, e$_4$=\textbf{D}efinition, e$_5$=\textbf{S}ubsumption, e$_6$=\textbf{L}egal \textbf{C}laim, e$_7$=\textbf{P}remise, with e$_3$ remaining unused throughout all data.
    We choose the joint task as it allows us to investigate the overall error of all components, rather than relying on the results of the previous stage or re-examining all components.
    However, we also report the results of our pre-trained LLM approach in the two-tiered setting as a point of comparison.

    In contrast to the AES dataset used in \cite{rodriguez2019language} and \cite{ormerod2021automated}, the GSHA dataset does not contain scores from multiple annotators.
    We therefore substitute the Quadratic Weighted Kappa (QWK) with linear and rank correlation and accuracy for evaluation.    

    \subsection{Gutachtenstil Argument Mining}\label{subsec:appraisal-style-argumentation-mining}

    Pre-trained LLMs cannot match the baseline extractive systems in SPWSLE \textit{Gutachtenstil} argument mining.
    In both the two-tiered extraction task (cf. table \ref{tab:spwsle_tt}) and the joint task (cf. table \ref{tab:spwsle_joint}) they fall behind the Bag-of-Words model and underperform especially for the subsumption categories.
    This can partially be explained by the entanglement of these categories and removing the relevant context, i.e. \textit{Gutachtenstil} explanation, counter-intuitively increases scores (c.f. table \ref{tab:spwsle_tt}, \texttt{Llama 3}$_{10}$ NE).
    This can be viewed as a manifestation of the inaccurate interpretation of legal concepts identified in \cite{lai2023large}.
    While Retrieval Augmented Generation helps performance by selecting the most relevant example shots, the overall delta in accuracy and F1 score is still significant at 18.1\% and 7.4\% respectively.

    However, we observe the BoW model to be on par with \citeposs{weber2023structured} BERT-based extractive classifier, offering a much more efficient mining system at equivalent effectiveness.

    \begin{table*}[ht]
    \footnotesize
        \centering
        \begin{tabular}{l||l|l|l|l|l||l|l|l}
            Method & \multicolumn{5}{c||}{Major Component F1} & \multicolumn{3}{|c}{Subsumption F1}  \\
                            & D & C & MC & S & N & P & LC & N \\\hline\hline
             BoW+SVM            &.\textbf{91}&.\textit{84}&.\textbf{96}&.\textit{65}&.\textit{77}&.\textit{67}&.\textit{48}&.\textbf{90}\\
             \texttt{BERT}$\dag$      &.82&.\textbf{89}&.\textit{93}&.\textbf{71}&.\textbf{88}&.\textbf{69}&.\textbf{66}&.\textit{78}\\
              \texttt{Llama 3}$_{10\text{RAG}}$   &.\textit{85}&.77&.85&.57&.45&.49&.27&.47\\
             \texttt{Llama 3}$_{10}$   &.{84}&.68&.71&.46&.44&.0&.0&.0\\
             \texttt{Llama 3}$_{10}$ NE&.75&.59&.51&.39&.36&.47&.16&.18
        \end{tabular}
        \caption{SPWSLE \textit{Gutachtenstil} argument mining results in the two-tiered task on a 20\% test set with \textbf{best} and \textit{second-best} results highlighted. For generative LLMs the subscript denotes the number of example shots provided in the prompt, \textit{RAG} = Retrieval Augmented Generation for example selection, \textit{NE} = No Explanation of the Gutachtenstil, $\dag$ = result from \cite{weber2023structured}.}
        \label{tab:spwsle_tt}
    \end{table*}  

    \begin{table}[ht]
        \centering
        \footnotesize
        \begin{tabular}{l||r|r}
            Method & Macro F1 & Accuracy   \\\hline\hline
             BoW+SVM            &.\textbf{761}&.\textbf{784}\\
             \texttt{Llama 3}$_{10\text{RAG}}$   &.\textit{580}&.\textit{710}\\
             \texttt{Llama 3}$_{10}$  &.{470}&.{585}\\
             \texttt{Mixtral}$_{10}$ &.370&.444\\
             \texttt{GPT-3.5}$_{10}$ &.339&.381
        \end{tabular}
        \caption{SPWSLE \textit{Gutachtenstil} joint task argument mining results on a 20\% test set.  For pre-trained LLMs the subscript denotes the number of examples provided in the prompt, \textit{RAG} = Retrieval Augmented Generation for example selection. Detailed results in \autoref{tab:spwsle_joint_detail}.}
        \label{tab:spwsle_joint}
    \end{table}

    \subsubsection{Retrieval Augmented Generation}

    To investigate the influence of RAG example selection, we perform a brittleness test by comparing the best and worst possible selection under the chosen criterion (cf. Table \ref{tab:spwsle_rag}).
    We observe this maximum performance margin to be 14.4\% macro F1 and 17.1\% accuracy overall with major regressions for the \textit{Major Claim} and \textit{Legal Claim} categories.
    With inverse RAG, providing the supposed least relevant 10 shots decreases performance below that of random example selection with a delta of 3.4\% macro F1 and 4.6\% accuracy.
    This hints that there are closely related or otherwise helpful samples in the training set, despite the disjointed documents, and that cosine similarity over embeddings can identify them.
    The retrieved example selection is not perfectly homogeneous, cf. figure \ref{fig:aes_rag}, so there is still reasoning required from the model.
    
    The worst-case selection of shots is still helpful in prompting, with a delta of 16.8\% F1 and 23.9\% accuracy over zero-shot prompting.
    These results confirm the overall benefit of RAG-based example selection and show that the penalty for sub-optimal selection is low.

    \subsubsection{Chain-of-Thought Prompting}

    The CoT strategy from \cite{wei2022chain} is not natively applicable to our few-shot prompting approach.
    We find that, if the model is presented with sample responses that do not also explain their reasoning, it will fail to do so for the current request as well.
    There are not detailed explanations available for SPWSLE, and while they could be generated for all training data, this would lead to a very expensive cold-start problem.
    As a compromise, we 
    limit the generation of artificial reasoning data to a small portion of the available training data.
    This ensures that examples are provided to the model with sufficient quality and diversity, while remaining tractable.
    We further investigate whether the advantages of CoT prompting can outweigh the lack of example prompts by implementing a CoT zero-shot model.
    
    \begin{table}[ht]
        \centering
        \footnotesize
        \begin{tabular}{l||r|r}
            Method & Macro F1 & Accuracy  \\\hline\hline
             CoT$_{10}^{80\%}$   &.\textbf{498}&.541\\
             Res$_{10}^{80\%}$      &.{470}&.\textbf{585}\\\hline
             Res$_{10}^{.03\%}$&   .\textbf{465}&.\textbf{587}\\
             CoT$_{10}^{.03\%}$   &.453&.484\\
             CoT$_{10\text{GAR}}^{.03\%}$&  .425&.470\\\hline
             CoT$_{0}$   &.\textbf{333}&.\textbf{393}\\
             Res$_{0}$    &.268&.30\\
        \end{tabular}
        \caption{\texttt{Llama 3} Chain-of-Thought prompting for SPWSLE \textit{Gutachtenstil} argument mining, with best-in-class results highlighted. Subscripts denote the number of shots provided in the prompt, superscripts indicate the training data available (80\%, 0.3\%, 0\%), \textit{Res} = result only prompt, \textit{CoT} = Chain-of-Thought prompt, \textit{GAR} = Generated Artificial Reasoning for prompt examples. Detailed results in \autoref{tab:spwsle_cot_detail}.}
        \label{tab:spwsle_cot}
    \end{table}

We find that with CoT prompting, the \texttt{Llama 3} shows some deficiencies  following instructions.
It is necessary to enforce an output format that ends with the category name to provide the explanation context before the final conclusion.
This routinely leads to malformed responses with fictional categories and, despite the explicit target language instruction (c.f. prompt \ref{prompt:cot}), in English instead of German.
With increased distance between the instruction and result, attention naturally lowers with the positional encodings diverging.
The effect that this has on instruction following outweighs any benefit of the more complex reasoning afforded by CoT prompting.

We therefore employ pattern matching and report the most frequent category in the model answer as the result.





While the generated artificial reasoning shots are generally of high quality, they do not improve consistency.
Effectively, this shows that there is error in the chain from feedback request to feedback to result.
Therefore either the models capability in providing feedback or integrating feedback is not no par, which we view as a critical issue in education use.

    \subsubsection{Pseudonymization}

    With generative pre-trained models, providing new context is difficult.
    Especially in the legal domain, certain terms carry significant meaning that cannot adequately be expressed in a short definition.
    In the case of \textit{Gutachtenstil} argument mining, we can check the presence of such meaningful internal representations in two ways.
    Firstly, we remove the \textit{Gutachtenstil} explanation from the system prompt, relying on knowledge acquired at training time and through example shots alone, and find that it only has a small impact on performance (cf. table \ref{tab:spwsle_tt} \texttt{Llama 3}$_{10}$ NE).
    
    However, we can effectively remove internal information about the categories by pseudonymizing them via random permutation for each query to the model.
    With pseudonymization the model cannot recover any meaningful information, cf. table \ref{tab:spwsle_pseudo}, confirming that this internalized knowledge is immediately necessary.

    \begin{table}[ht]
        \centering
        \footnotesize
        \begin{tabular}{l||r|r}
            Labels & Macro F1 & Accuracy \\\hline\hline
             Original   &.\textbf{470}&.\textbf{585}\\
             Pseudonyms &.134&.159\\
        \end{tabular}
        \caption{Inherent knowledge test via pseudonymization for \texttt{Llama 3}$_{10}$ in the SPWSLE \textit{Gutachtenstil} argument mining joint task. Labels are either original or randomly permuted for each request. Detailed results in \autoref{tab:spwsle_pseudo_detail}.}
        \label{tab:spwsle_pseudo}
    \end{table}

\subsection{Public Feedback Argument Mining}

    \begin{table}[ht]
        \centering
        \footnotesize
        \begin{tabular}{l||r|r}
            Model & Macro F1& Accuracy \\\hline\hline
            SVM $\dag$            &.\textbf{73}$_{\pm.02}$&\\
            BoW+SVM         &.{526}$_{\pm.02}$& .\textit{743}$_{\pm.01}$\\\hline
            \texttt{Llama 3}$_0$      &.445$_{\pm.01}$&.612$_{\pm.02}$\\
            \texttt{Llama 3}$_0$CoT   &.457$_{\pm.03}$&.627$_{\pm.02}$\\
            \texttt{Llama 3}$_{10}$   &.505$_{\pm.02}$&.{733}$_{\pm.02}$\\
            \texttt{Llama 3}$_{10}$CoT&.503$_{\pm.01}$&.732$_{\pm.02}$\\
            \texttt{Llama 3}$_{10\text{RAG}}$&.\textit{616}$_{\pm.03}$&.\textbf{779}$_{\pm.02}$\\
        \end{tabular}
        \caption{Argument mining scores for the CIMT CD\_C set in 3-fold cross validation. $\dag$ = results for SVM with unspecified features from \cite{romberg2021citizen}.}
        \label{tab:cimt}
    \end{table}

As the CIMT dataset contains only 4 categories and their structure is clearer than SPWSLE, we find performance improved generally, cf. table \ref{tab:cimt}.
The features used for the SVM in \cite{romberg2021citizen} are not specified, so we were not able to reproduce their results.
However, \texttt{Llama 3} can be made competitive with the BoW+SVM baseline and even surpass it with 10 example shots curated via RAG.
Again we find that CoT prompting is beneficial for the zero-shot case, however only marginally, but does not offer much, if any, improvement in few-shot prompting.

\subsection{Legal Essay Scoring}\label{subsec:legal-case-solution-grading}

Our selection of pre-trained LLMs does not perform well in legal essay scoring.
Even though their context windows are large enough to include up to 6 examples, they cannot achieve prediction quality beyond random guessing (cf. fig. \ref{fig:aes_shots}). 
Various techniques to lower task difficulty, such as simpler grading, additive scoring scale, partial scoring and chain-of-thought prompting did not yield any improvement.
We surmise that this grading task is too complex and noisy to be solved with pre-trained generative models of this calibre.

    \subsection{Student Essay Scoring}

    The Automated Student Assessment Prize dataset represents a set of much easier scoring tasks than the GSHA.
    As the texts are both more plentiful and generally much shorter, long contexts and instruction following are less problematic.
    Accordingly, \texttt{Llama 3} with 10 example shots outperforms the BoW baseline here with considerable margin, cf. table \ref{tab:asap}.
    However the zero-shot scenario regresses considerably in linear correlation and CoT does not offer any improvement.
    This suggests that even with the slightly more complex set 8 we are observing the limits of model capability.
    
    The improvements over the baseline with \texttt{Llama 3}$_{10}$ are still a good sign for LLM-based AES in education settings.

    \begin{table}[ht]
        \centering
        \footnotesize
        \begin{tabular}{l||r|r}
            Model & Spearman & Pearson\\\hline\hline
            BoW+SVM             & .{586}$_{\pm.04}$&.{616}$_{\pm.07}$\\\hline
            \texttt{Llama 3}$_0$      &.50$_{\pm.02}$&.405$_{\pm.08}$\\
            \texttt{Llama 3}$_{0}$CoT  &.582$_{\pm.03}$&.614$_{\pm.01}$\\
            \texttt{Llama 3}$_{10}$     &.\textit{643}$_{\pm.05}$&.\textbf{664}$_{\pm.03}$\\
            \texttt{Llama 3}$_{10}$CoT  &.539$_{\pm.04}$&.442$_{\pm.05}$\\
            \texttt{Llama 3}$_{10\text{RAG}}$     &.\textbf{658}$_{\pm.05}$&.\textit{658}$_{\pm.03}$\\
        \end{tabular}
        \caption{AES correlation scores for ASAP set 8 in 3-fold cross validation.}
        \label{tab:asap}
    \end{table}

    The main advantage of pre-trained generative LLMs is their adaptability to new tasks, which we test in a transfer study.
    In a zero-shot scenario this requires no new training data be collected for the new task, while in few-shot prompting example shots from another task can still be useful.
    The baseline and LLM models operate here on training data, be it for example shots or from-scratch fitting, from one set for the prediction on another set.

    \begin{figure}
        \centering
        \begin{subfigure}{.49\linewidth}
            \includegraphics[width=\textwidth]{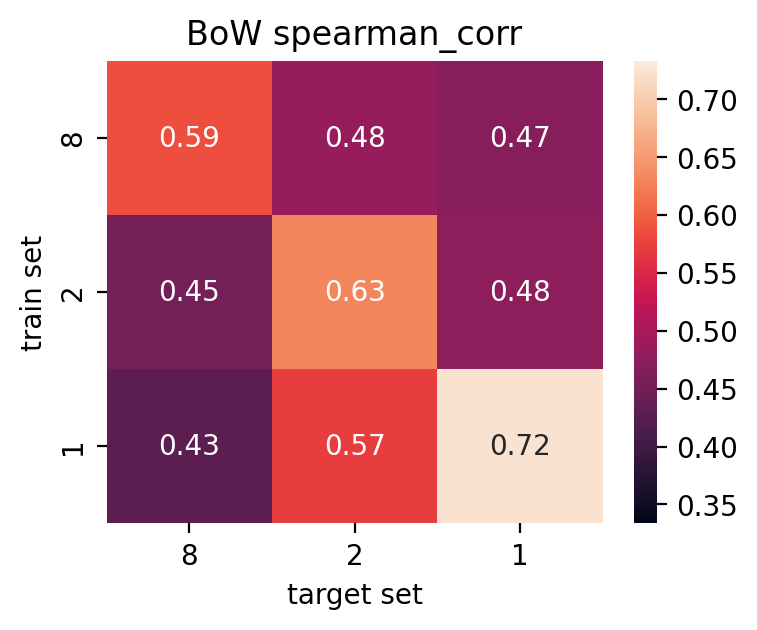}
            \caption{BoW}
        \end{subfigure}
        \begin{subfigure}{.49\linewidth}
            \includegraphics[width=\textwidth]{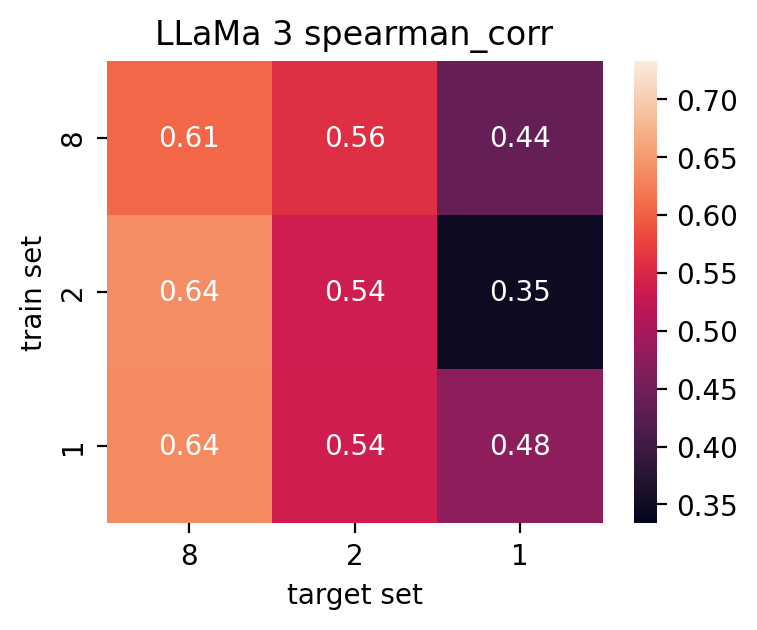}
            \caption{\texttt{Llama 3}$_{10}$}
        \end{subfigure}

        \caption{Spearman ranking correlation for the ASAP AES transfer study, mean over 3 folds.}
        \label{fig:asap_transfer}
    \end{figure}

As per the transfer study results (cf. figure \ref{fig:asap_transfer}), the penalty for providing foreign example shots to the model is low, and they still offer improved performance over zero-shot.
While the baseline model suffers from a domain shift due to its inflexible vocabulary and representations, \texttt{Llama 3}$_{10}$ adapts to sets 8 and 2 without issues.
We believe that this is rooted in the fact that examples are mostly useful for instruction following and their actual content and exact response are secondary.
This confirms our earlier results that show the negligible impact of sub-optimal example shot selection.

    \section{Conclusion}\label{sec:conclusion}

We identified that instruction following and especially German language understanding, are key factors for a significant gap between LLM capabilities and the demands of German legal education. 
The models studied fail to demonstrate the logical structuring needed for legal understanding and thus cannot effectively score full legal essays. 
However, simpler English tasks show more promising results, suggesting that pre-trained LLMs could be useful with basic tasks or more advanced models.

Our findings indicate that \texttt{Llama 3}'s superior language comprehension trumps \texttt{GPT-3.5}'s reasoning capabilities, highlighting the importance of linguistic skills in natural language communication.
Most models benefit from few-shot prompting, while Chain-of-Thought prompting only helps in zero-shot scenarios. 
The failure of our Auto-CoT inspired approach providing generated reasoning before extracting final results shows a lack of feedback integration capability.

We identified certain limitations such as inference efficiency and target language understanding in the current generation of LLMs.
Even with the efficiency gains provided by Grouped-Query Attention and Mixture-of-Expert models, we find that for incremental tasks, this makes them prohibitively expensive.
For languages outside of the main training language, even with multilingual models, proficiency is still lacking.

Overall, prompt-based extractive use of LLMs works well for simple tasks but underperforms in complex tasks with long contexts due to limited language understanding and instruction following. 
This can be partially mitigated with careful selection of examples.

     \paragraph{Limitations}\label{subsec:limitations}

     While pre-trained LLMs lift the large burden of training, the inference compute is still expensive, which cannot be solved with efficient prompting or fine-tuning alone.

     As the examined models benefit from abundant training data, knowledge transfer to new within-domain tasks should be checked to gauge the efficiency of adaptation.
     However, it was not possible for us to conduct such a study as the linking to the case is not included in the documents or annotations.

     As shown by much improved results in English language AES tasks, even with multilingual models, the proficiency in any but the main training language is still lacking.

    \bibliography{main}

    \appendix

    \section{Appendix}
    \label{sec:appendix}

    \begin{table*}[ht]
    \footnotesize
        \centering
        \begin{tabular}{l||r|r|r}
             Model& Global&Reasoning&Language \\\hline
             meta-llama-3-70b-instruct& \textbf{35.99}&22.67&\textbf{34.11}\\
             gpt-3.5-turbo-0125& \textit{34.77}&\textbf{28.00}&\textit{24.22}\\
             mixtral-8x7b-instruct-v0.1& 23.39&\textit{23.33}&13.76\\
        \end{tabular}
        \caption{\normalsize \textit{LiveBench} \cite{livebench} LLM accuracy scores (\%) for categories and models relevant to this work.}
        \label{tab:livebench}
    \end{table*}

        \prompt{
    Annotate texts according to the Gutachtenstil.

$<$EXPLANATION$>$

The text must be assigned to exactly one of the following categories: “Major Claim”, “Conclusion”, “Definition”, “Subsumption”, “Legal Claim”, “Premise” or “Unknown”

Answer in one word.

Your answer should only mention the relevant component.
    }{Translated system prompt for SPWSLE \textit{Gutachtenstil} argument mining where $<$EXPLANATION$>$ defines the Gutachtenstil.}{prompt:spwsle_sys}

    \prompt{
        Answer in one word. Which part of the Gutachtenstil is this?
        }{Translated result prompt for SPWSLE \textit{Gutachtenstil} argument mining.}{prompt:spwsle}

    \begin{table*}[ht]
        \centering
        \footnotesize
        \begin{tabular}{l||r|r||r|r|r|r|r|r}
            Method & Macro & Acc. & \multicolumn{6}{c}{Class F1}  \\
                            &  F1&    & D & C &MC & P & LC& S \\\hline\hline
             BoW+SVM            &.\textbf{761}&.\textbf{784}&.\textbf{91}&.\textbf{84}&.\textbf{96}&.\textbf{67}&.\textbf{54}&.\textbf{65}\\
             \texttt{Llama 3}$_{10\text{RAG}}$   &.\textit{580}&.\textit{710}&.\textit{89}&.\textit{79}&.\textit{86}&.\textit{62}&.\textit{33}&.\textit{56}\\
             \texttt{Llama 3}$_{10}$  &.{470}&.{585}&.{85}&.{67}&.{68}&.{54}&.07&.{47}\\
             \texttt{Mixtral}$_{10}$ &.370&.444&.73&.48&.54&.42&.09&.34\\
             \texttt{GPT-3.5}$_{10}$ &.339&.381&.63&.45&.47&.31&.{17}&.35
        \end{tabular}
        \caption{Detailed results for SPWSLE \textit{Gutachtenstil} joint task argument mining on a 20\% test set.  For pre-trained LLMs the subscript denotes the number of examples provided in the prompt, \textit{RAG} = Retrieval Augmented Generation for example selection.}
        \label{tab:spwsle_joint_detail}
    \end{table*}

    \begin{table*}[ht]
        \centering
        \footnotesize
        \begin{tabular}{l||r|r||r|r|r|r|r|r}
            Method & F1 & Acc. & \multicolumn{6}{c}{Class F1}  \\
                            &  &    & D & C &MC & P & LC& S \\\hline\hline
             \texttt{Llama 3}$_{10\text{RAG}}$   &.\textbf{580}&.\textbf{710}&.\textbf{89}&.\textbf{79}&.\textbf{86}&.\textbf{62}&.\textbf{33}&.\textbf{56}\\
             \texttt{Llama 3}$_{10\overline{\text{RAG}}}$   &.{436}&.{539}&.{82}&.{64}&.{53}&.{53}&.{11}&.{42}\\\hline
             $\Delta$   &.{144}&.{171}&.{07}&.{15}&.{33}&.{09}&.{22}&.{14}\\
        \end{tabular}
        \caption{RAG Brittleness test for SPWSLE \textit{Gutachtenstil} argument mining on a 20\% test set. $\overline{\text{RAG}}$ = Inverse Retrieval Augmented Generation for example selection.}
        \label{tab:spwsle_rag}
    \end{table*}

    \begin{table*}[ht]
        \centering
        \footnotesize
        \begin{tabular}{l||r|r||r|r|r|r|r|r}
            Method & Macro & Acc. & \multicolumn{6}{c}{Class F1}  \\
                            &  F1&    & D & C &MC & P & LC& S \\\hline\hline
             CoT$_{10}^{80\%}$   &.\textbf{498}&.541&.72&.65&.68&.53&.\textbf{09}&.32\\
             Res$_{10}^{80\%}$      &.{470}&.\textbf{585}&.\textbf{85}&.\textbf{67}&.\textbf{68}&.\textbf{54}&.07&.\textbf{47}\\\hline
             Res$_{10}^{.03\%}$&   .\textbf{465}&.\textbf{587}&.85&.\textbf{68}&.\textbf{70}&.\textbf{51}&.03&.\textbf{48}\\
             CoT$_{10}^{.03\%}$   &.453&.484&.\textbf{86}&.60&.33&.49&.\textbf{08}&.35\\
             CoT$_{10\text{GAR}}^{.03\%}$&  .425&.470&.82&.60&.32&.53&.01&.26\\\hline
             CoT$_{0}$   &.\textbf{333}&.\textbf{393}&.57&.\textbf{51}&.45&.\textbf{29}&.\textbf{05}&.13\\
             Res$_{0}$    &.268&.30&.\textbf{63}&.29&.\textbf{50}&.04&.05&.\textbf{37}\\
        \end{tabular}
        \caption{Detauled results for \texttt{Llama 3} Chain-of-Thought prompting in different data availability scenarios (80\%, 0.3\%, 0\%) for SPWSLE \textit{Gutachtenstil} argument mining, with best-in-class results highlighted. The subscript denotes the number of shots provided in the prompt, the superscript indicates the training data available if applicable, \textit{Res} = result only prompt, \textit{CoT} = Chain-of-Thought prompt, \textit{GAR} = Generated Artificial Reasoning for prompt examples.}
        \label{tab:spwsle_cot_detail}
    \end{table*}

    \prompt{
        Text:
        
$<$TEXT$>$

Explain: What part of the Gutachtenstil is this?
Briefly explain your decision in German, up to 100 words.
}{Chain-of-Thought prompt for SPWSLE \textit{Gutachtenstil} argument mining}{prompt:cot}
    
    \begin{figure}[ht]
        \centering
        \includegraphics[width=\linewidth]{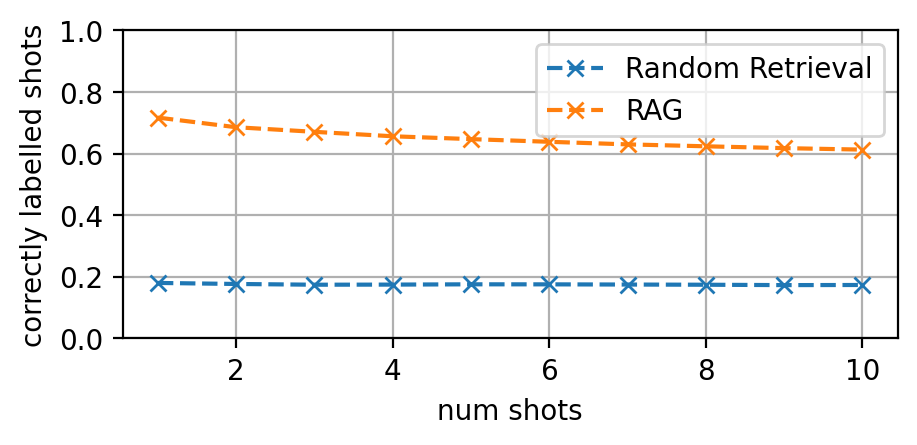}
        \caption{Mean proportion of shots with the correct label for a request in SPWSLE \textit{Gutachtenstil} argument mining. Random selection vs. RAG-based selection via cosine similarity over sentence embeddings.}
        \label{fig:aes_rag}
    \end{figure}

    \begin{table*}[ht]
        \centering
        \footnotesize
        \begin{tabular}{l||r|r||r|r|r|r|r|r}
            Labels & Macro & Acc. & \multicolumn{6}{c}{Class F1}  \\
                            &  F1&    & D & C &MC & P & LC& S \\\hline\hline
             Original   &.\textbf{470}&.\textbf{585}&.\textbf{85}&.\textbf{67}&.\textbf{68}&.\textbf{54}&.07&.\textbf{47}\\
             Pseudonyms &.134&.159&.12&.18&.20&.16&\textbf{.13}&.14\\
        \end{tabular}
        \caption{Detailed results for inherent knowledge test via pseudonymization for \texttt{Llama 3}$_{10}$ in the SPWSLE \textit{Gutachtenstil} argument mining joint task. The labels are either original or randomly permuted for each request.}
        \label{tab:spwsle_pseudo_detail}
    \end{table*}

    \begin{figure}[ht]
        \centering
        \includegraphics[width=\linewidth]{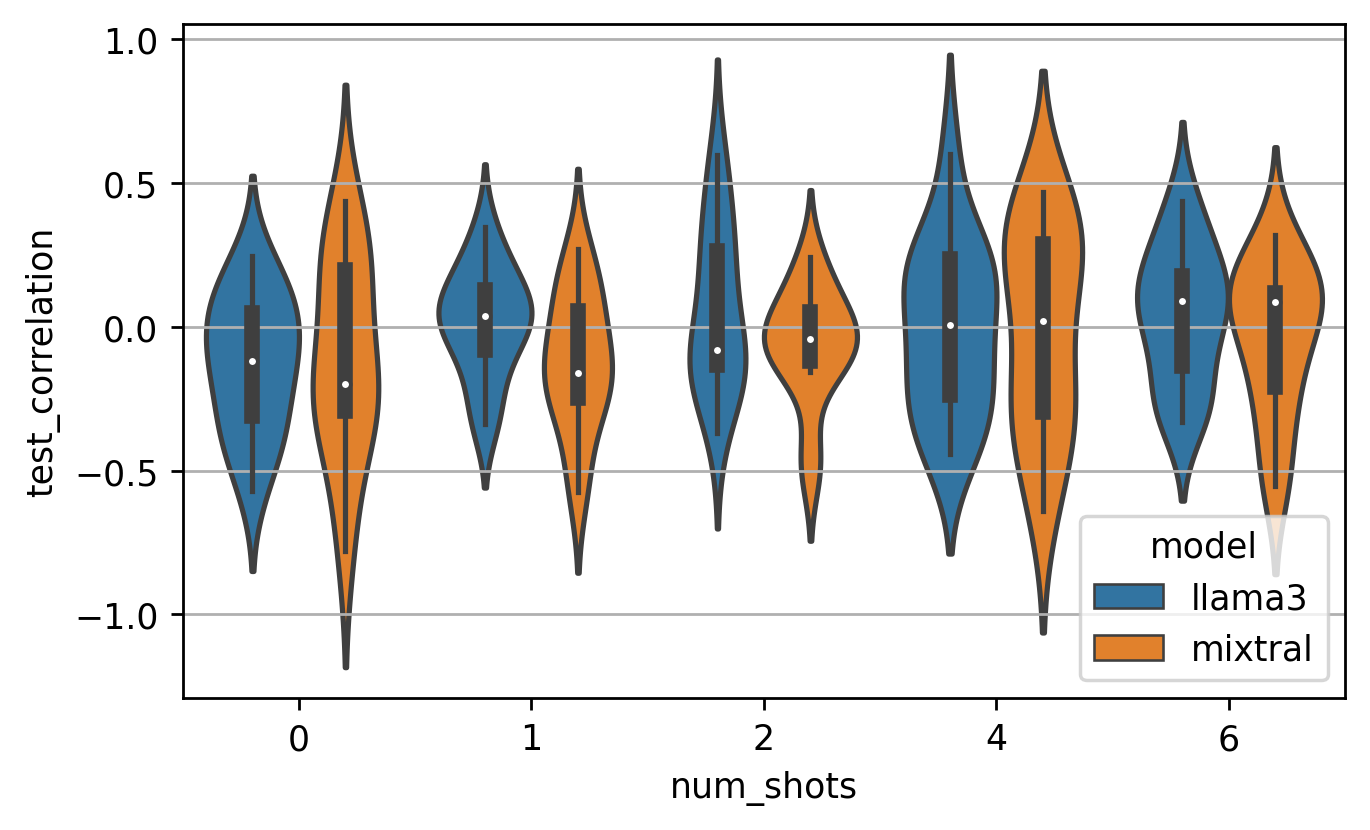}
        \caption{GSHA automatic essay scoring results with few-shot prompting in 5-fold cross-validation repeated 3 times.
        }
        \label{fig:aes_shots}
    \end{figure}

\end{document}